\documentclass[10pt,twocolumn,letterpaper]{article}

\usepackage{wacv}
\usepackage{times}
\usepackage{epsfig}
\usepackage{graphicx}
\usepackage{subfigure}
\usepackage{amsmath}
\usepackage{amssymb}

\renewcommand{\vec}[1]{\mathbf{#1}}

\wacvfinalcopy % *** Uncomment this line for the final submission

\def\wacvPaperID{344} % *** Enter the wacv Paper ID here

\ifwacvfinal\fi
\begin{document}

%%%%%%%%% TITLE
\title{SHADHO: Massively Scalable Hardware-Aware Distributed Hyperparameter Optimization}

% Authors at the same institution
\author{Jeffery Kinnison \hspace{1cm} Nathaniel Kremer-Herman \hspace{1cm} Douglas Thain \hspace{1cm} Walter Scheirer  \\
University of Notre Dame\\
{\tt\small \{jkinniso,nkremerh,dthain,walter.scheirer\}@nd.edu}
}

\maketitle
\ifwacvfinal\thispagestyle{empty}\fi

%%%%%%%%% ABSTRACT
\begin{abstract}
   Computer vision is experiencing an AI renaissance, in which machine learning models are expediting important breakthroughs in academic research and commercial applications. Effectively training these models, however, is not trivial due in part to hyperparameters: user-configured values that control a model's ability to learn from data. Existing hyperparameter optimization methods are highly parallel but make no effort to balance the search across heterogeneous hardware or to prioritize searching high-impact spaces. In this paper, we introduce a framework for massively Scalable Hardware-Aware Distributed Hyperparameter Optimization (SHADHO). Our framework calculates the relative complexity of each search space and monitors performance on the learning task over all trials. These metrics are then used as heuristics to assign hyperparameters to distributed workers based on their hardware. We first demonstrate that our framework achieves double the throughput of a standard distributed hyperparameter optimization framework by optimizing SVM for MNIST using 150 distributed workers. We then conduct model search with SHADHO over the course of one week using 74 GPUs across two compute clusters to optimize U-Net for a cell segmentation task, discovering 515 models that achieve a lower validation loss than standard U-Net.
\end{abstract}
\vspace{-1em}

%%%%%%%%% BODY TEXT
\section{Introduction}

Without question, advances in high-performance computing architectures have fueled the innovation and success of computationally expensive machine learning methods. New many-core architectures allow researchers to make full use of the large labeled datasets that are necessary to train effective models (i.e., solutions) for various learning problems. Modern machine learning tools backed by GPUs, clusters of traditional CPUs, and custom parallel hardware have enabled data-driven computer vision approaches in numerous domains, including healthcare~\cite{esteva2017dermatologist}, autonomous vehicles~\cite{DBLP:journals/corr/BojarskiTDFFGJM16}, human biometrics~\cite{Parkhi15}, and many more~\cite{fergus2013}.

Despite these successes, selecting the correct model for particular data remains a difficult problem. Model performance is highly algorithm-specific, with different models producing wildly different results on the same dataset. However, model selection is not simply an algorithmic choice. Model searches must also account for hyperparameters: free parameters associated with a particular machine learning model that govern its ability to learn. These parameters are separate from the elementary parameters (i.e., weights) that are learned from the data, and are set before training takes place. Hyperparameters are often defined over nonlinear, non-convex spaces with many local minima, making optimization non-trivial. Choosing the best model for a particular learning task boils down to choosing the parametrized model that can accurately make predictions from new data. This is known as the \textit{hyperparameter optimization problem}.

\begin{figure*}[!htp]
    \centering
    \includegraphics[width=\textwidth]{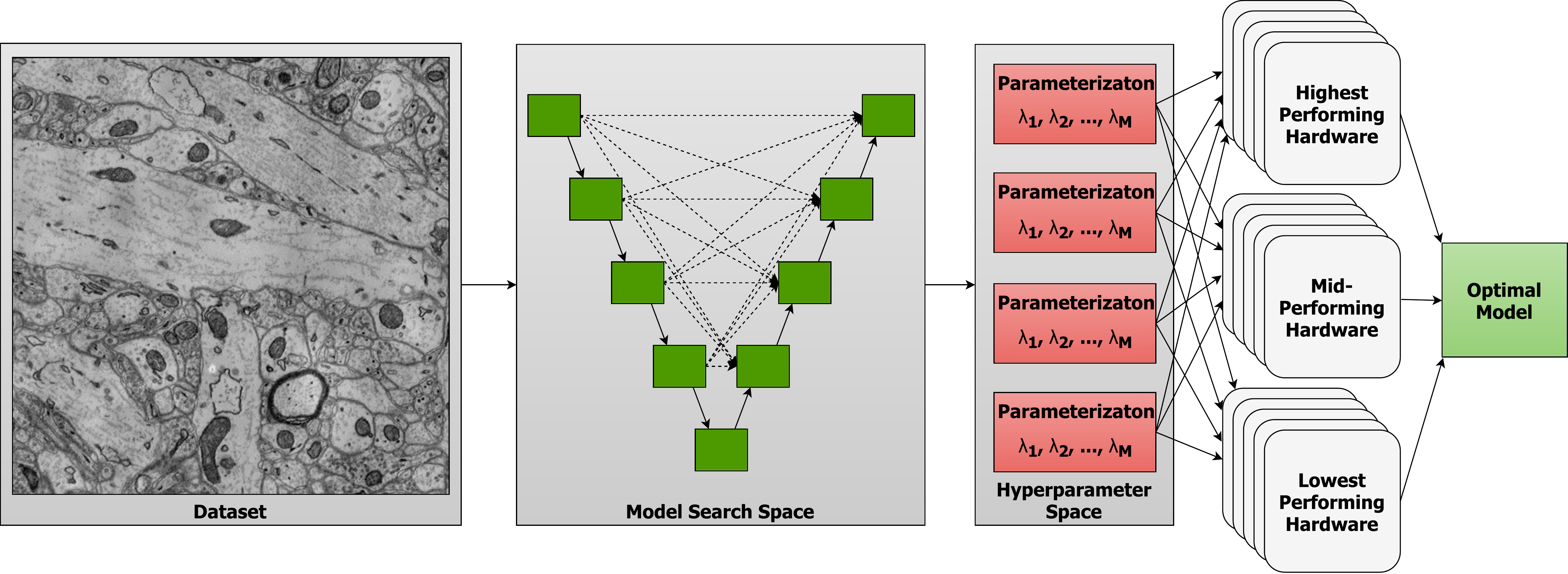}
    \caption{Example distributed hyperparameter optimization of U-Net~\cite{ronneberger-2015-unet} with different bypass connections for cell segmentation in images from electron microscopy~\cite{kasthuri-2015-cell}. To discover an optimal model, first a configuration of bypass connections is selected, then the layers of the network are parametrized. The parametrized model is then sent to a remote worker for training and evaluation. This process is repeated a large number of times, and the optimal parametrized model is returned. Existing distributed hyperparameter optimization frameworks assign a model to the first available worker. SHADHO, by contrast, uses model structure and search performance to influence search throughput by directing models to appropriate hardware.}
    \vspace{-1em}
    \label{fig:teaser}
\end{figure*}

Steps have been made toward local \cite{kotthoff-2016-autoweka,pedregosa-2011-scikit-learn} and distributed \cite{bergstra-2013-hyperopt,claesen-2014-optunity,dewancker-2016-sigopt,snoek-2012-spearmint,aetros-2016,h2oai-2015} hyperparameter search strategies. An example of the distributed hyperparameter optimization process is shown in Figure~\ref{fig:teaser}. Curiously, though, current distributed strategies do not take advantage of potential systems-based optimizations. Existing solutions do not account for the hardware being used, instead enforcing (in the case of cloud-based platforms \cite{dewancker-2016-sigopt,aetros-2016}) or assuming that all connected hardware is identical. A key problem that we and many other machine learning researchers have encountered is that available computing resources for distributed hyperparameter optimization are typically heterogeneous and often spread across different networks. This is the state of affairs for nearly all users outside of a handful of cloud-based providers, which have internal on-demand access to large homogeneous collections of fast hardware. In the former case, efficient hyperparameter searches must adjust the search strategy to make the best use of available hardware. Even in the latter case, though, for collections of homogeneous %multi- and many-core
systems, decisions may be made about the level of parallelism to exploit during the search.

While adapting distributed hardware to a search is not possible, information about the search can be used to adapt it to the hardware. Hyperparameters are ubiquitous across different classes of machine learning algorithms, and each algorithm can have a different number of hyperparameters distributed over different numeric and categorical spaces. For example, a Support Vector Machine (SVM) can have one to four numeric parameters depending on the kernel function used (i.e., linear,  radial basis function, sigmoid, and polynomial). Turning to deep learning, convolutional neural networks (CNNs) require tuning a number of hyperparameters within each convolutional layer, meaning that the search must expand with the size of the network. Information about the structure of the search, such as the hyperparameter count of each model, can inform how optimization proceeds.

% For example, a linear Support Vector Machine (SVM) model has a single hyperparameter (the soft-margin constant) that is incremented logarithmically within a fixed range, a Radial Basis Function SVM has an additional hyperparameter (the kernel coefficient) defined over an infinite real-valued space, a sigmoid SVM has yet another infinitely ranged hyperparameter, and a polynomial SVM has a fourth that controls the degree of the polynomial. In these examples, the search over polynomial SVM hyperparameters is combinatorially larger than that of the linear SVM, with more possible search values. Turning to deep learning, convolutional neural networks (CNNs) require tuning of a number of hyperparameters within each convolutional layer, meaning that the search must expand with the size of the network.

A number of hyperparameter search strategies have been proposed to solve this optimization problem, ranging from naive approaches like random search \cite{bergstra-2012-random-search}, to more rigorous approaches like Bayesian optimization \cite{bergstra-2011-tpe,snoek-2012-spearmint}, gradient-based learning \cite{maclaurin-2015-gradient}, and bandit-based searches \cite{li-2016-bandit}. These strategies address the problem of choosing the next parametrization to test, but they operate under two major simplifying assumptions: 1) that hyperparameters have equal priority in the search, and 2) that hyperparameter search spaces are equally complex. In practice, neither of these assumptions are generally true.

One avenue for improving distributed hyperparameter optimization, then, is to account for the necessity and complexity of searching hyperparameter spaces to improve scheduling the search. Hyperparameter optimization is best represented as a bag of tasks (BoT) application, in which there are an effectively infinite number of tasks that must be mapped to a finite set of resources and no task is explicitly dependent upon another. Existing hyperparameter optimization software uses a naive first-come, first-serve (FCFS) approach to schedule hyperparameter evaluations. By incorporating information about the models and hyperparameters being searched, however, it is possible to schedule such that a greater proportion of searches are allocated to larger models with less certain performance.

In this paper, we present the Scalable Hardware-Aware Distributed Hyperparameter Optimization framework (SHADHO), a general-purpose hyperparameter optimization framework. For each model in the search, SHADHO approximates the complexity (aggregate size of the hyperparameter domains) and priority (variation in performance across different parametrizations). Models are ranked by these two heuristics, and high-complexity / high-priority models are assigned to more performant hardware. In this way, search throughput is increased across models with many hyperparameters and wider ranges of performance on the learning task, making SHADHO suitable for applications such as neural network architecture search. In summary, the contributions of this paper are as follows
\vspace{-0.5em}
\begin{enumerate}
    \setlength\itemsep{-0.5em}
    \item A comprehensive survey of existing hyperparameter optimization methods.
    \item Two heuristics --- complexity and priority --- for ranking models in a model search / hyperparameter optimization process.
    \item A description of the SHADHO framework.
    \item A demonstration of what the increased throughput heuristic-based scheduling with SHADHO can offer over FCFS schemes.
    \item An application of SHADHO to CNN optimization for membrane detection in microscopic images.
\end{enumerate}

\section{Related Work}

\textbf{Hyperparameter Optimization.} Hyperparameter optimization methods typically address the problem of how to choose the next hyperparameter value to search. Manual tuning and grid search~\cite{duan-2005-grid} are popular methods because they are easy to implement, however they rely upon domain knowledge and will skip over many values in continuous domains. Bergstra and Bengio~\cite{bergstra-2012-random-search} argued to replace these practices with random search because it is just as easy to implement and does not require discretizing the search space. This scheme will more thoroughly search the hyperparameter spaces, however it has no mechanism for narrowing the scope of the search. Random search is the basis for Hyperopt~\cite{bergstra-2015-hyperopt,bergstra-2013-hyperopt}, a widely-used open source hyperparameter optimization framework.

Guided approaches to hyperparameter optimization are also popular, notably genetic algorithms and Bayesian optimization strategies. Genetic algorithms~\cite{benardo-2007-optimizing,friedrichs-2005-evolutionary,vose-2017-genetic-algorithm} have historically been applied to hyperparameter optimization, however they are prohibitively expensive as the number of hyperparameters increases~\cite{bergstra-2011-tpe}. Sequential Model-based Bayesian Optimization~\cite{jones-1998-smbo}, Tree-Structured Parzen Estimators~\cite{bergstra-2011-tpe}, Gaussian process-based estimation~\cite{snoek-2012-spearmint}, and Sequential Model-based Algorithm Configuration~\cite{hutter-2013-smac} are popular Bayesian optimization strategies, implemented in a number of open source and proprietary frameworks~\cite{bergstra-2013-hyperopt,claesen-2014-optunity,dewancker-2016-sigopt,kotthoff-2016-autoweka,snoek-2012-spearmint,aetros-2016,h2oai-2015}. These methods use previous hyperparameter values and their corresponding evaluations as priors for approximating viable hyperparameter values, and as such can become stuck in local minima.

Methods beyond genetic algorithms and Bayesian optimization have also been explored to exploit different selection criteria. MacLaurin \textit{et al.}~\cite{maclaurin-2015-gradient} introduced a method that learns gradients with respect to hyperparameter values, allowing for fine-grained hyperparameter optimization but requiring an expensive gradient calculation step. Domhan \textit{et al.}~\cite{domhan-2015-learning-curve} introduced a method for extrapolating learning curves using Markov-Chain Monte Carlo inference to predict parametrized model performance and update the hyperparameter selection method using the predicted performance. The Hyperband method introduced by Li \textit{et al.}~\cite{li-2016-bandit} performs grid search using a budgeted successive halving method that assigns a cost to each search. Ilievski \textit{et al.}~\cite{ilievski-2017-rbf} propose using radial basis function surrogates and dynamic coordinate search to select candidate hyperparameter values. Like Bayesian optimization, these each narrow the viable domain of each hyperparameter under optimization by examining previously tested values and applying an operation to prune the domain. As with Bayesian optimization, these methods run the risk of falling into local minima.

Tuning a neural network architecture for a particular learning problem presents a different set of challenges to standard hyperparameter tuning, including selecting neural network layers and connections between layers. The optimal neural network architecture is typically determined by iteratively building up the network and observing performance on the dataset. Historically, this iterative procedure has been carried out by trial-and-error~\cite{fernandes-2005-trial-and-error,furtuna-2011-optimization}, in which one parameter from one layer is manually varied at a time. Several automated methods, including construction / pruning methods~\cite{hu-2016-pruning}, particle swarm optimization~\cite{garro-2015-particle-swarm}, and genetic algorithms~\cite{benardo-2007-optimizing} select new layers based on observed performance. Zoph \textit{et al.}~\cite{zoph-2016-archsearch,zoph2017learning} developed a method that chooses optimal neural network models using reinforcement learning, however they report using several hundred GPUs to conduct this search in both published case studies. Neural networks are also being applied on a smaller scale for architecture selection~\cite{brock2017smash,cortes-2016-adanet}, utilizing a network trained to generate candidate architectures for evaluation.

All of these hyperparameter optimization and architecture search methods were created to decide \textbf{which} hyperparameter values to search while assuming all models and hyperparameters are equal. The question of \textbf{how} to conduct a hyperparameter search given a set of possibly heterogeneous distributed resources, particularly in terms of directing models to appropriate hardware and focusing on models with uncertain performance, is new to this problem.

\textbf{Dynamic Distributed Task Scheduling.}
A number of heuristic-based distributed scheduling algorithms exist for general-purpose task scheduling, with a focus on optimizing for task dependency graphs. The two algorithms (HEFT and CPOP) presented in \cite{heft-2002} focus on providing solutions to scheduling tasks in a directed acyclic graph (DAG) structure to prioritize critical tasks which many other tasks rely upon. HEFT uses a task complexity heuristic to schedule dependent tasks similar to the heuristic we propose. In \cite{heft-lookahead-2010}, the authors built upon the HEFT algorithm presented in \cite{heft-2002} to more explicitly consider the effect scheduling certain tasks before others will have on the performance of the DAG workflow. Biswas \textit{et al.}~\cite{biswas-2017-multiqueue} extended the HEFT algorithm with additional heuristics and a multi-queue scheduler to address the problem of scheduling across heterogeneous systems. Al Ebrahim and Ahmad~\cite{alebrahim-2017-task-scheduling} also extended HEFT to explicitly consider all dependencies and data transfer among a static set of tasks.

Existing heuristic scheduling solutions are not well-suited for solving the problem of scheduling a model search or hyperparameter optimization process because they assume that the task space is modeled as a dependency graph. Hyperparameter optimization is best modeled as a BoT application: mapping an effectively infinite set of independent tasks to a finite set of resources. Heuristic scheduling for BoT typically involves monitoring resource utilization~\cite{zhang-2016-bot,zhang-2017-bot} to minimize the cost of running the tasks, or else scaling hardware to the set of tasks in the case of elastic cloud services~\cite{Silva:2008:HRA:1462704.1462713}. This work, by contrast, schedules based on properties of the tasks (i.e., model training sessions) themselves to match the available hardware, focusing on larger models with less certainty in their performance.
%Moreover, these methods often assume a static (i.e., finite) set of tasks to schedule. In many cases, model search involves a possibly infinite, completely independent distributed tasks, or else dependencies cannot be known \textit{a priori}, as in iterative refinement of the viable model spaces and hyperparameter domains. This work builds upon heuristic task scheduling work by tailoring the heuristics to the model search / hyperparameter optimization domain.

\section{Heuristics for Hyperparameter Search}

Fundamentally, hyperparameter optimization is the problem of selecting a model, $\mathcal{M}$, and parametrizing it with a set of hyperparameters, $\vec{\lambda}$, such that $\mathcal{M}(\vec{\lambda})$ learns a desired set of patterns from a dataset with minimal error. Each hyperparameter $\lambda_{i} \in \vec{\lambda}$ is defined over a distinct discrete or continuous domain, $s_{i} \in \vec{s}$, and hyperparameters are searched by drawing a value from each domain and evaluating the performance of $\mathcal{M}(\vec{\lambda})$.

Optimizing hyperparameter search over a particular set of hardware is the problem of mapping models to hardware with resources proportionate to the need to search a given model. To measure ``need," two pieces of information must be known: 1) the number of searches necessary to completely search $\mathcal{M}(\lambda)$, and 2) the fitness of $\mathcal{M}$ to the learning problem. Both of these values must be approximated in general because hyperparameters are often defined over continuous domains, and model performance may only be determined experimentally. We define these approximations as the search \textit{complexity} and \textit{priority} of each model.

\subsection{Complexity}

The complexity of a model is determined by the aggregate size of its viable hyperparameter domains. Hyperparameter domains are defined over both continuous and discrete spaces, so any heuristic must be an necessarily approximation of the size of the search. Moreover, a complexity heuristic should maintain the order of spaces based on their size. Thus, for any hyperparameter domain $s_{i} \in \vec{s}$, we define the search complexity $C(s)$ as
\begin{equation} \label{eqn:complexity}
C(s_{i}) =
\begin{cases}
2 + \|b - a\| &  \textit{if s is continuous} \\
2 - \frac{1}{|s_{i}|} & \textit{if s is discrete}
\end{cases}
\end{equation}

\noindent
where $[a, b]$ is the closed interval containing $99\%$ of the probability distribution governing $s$. $C(s)$ enforces a strong ordering on spaces based on their size: continuous spaces are considered more complex to search than discrete spaces. Moreover, $C(s)$ maintains the order of continuous spaces relative to continuous spaces and discrete spaces relative to discrete spaces. Model complexity is then approximated as $C(\vec{s}) = \sum_{s_{i} \in \vec{s}}C(s_{i})$.

\begin{figure*}[!htp]
    \centering
    \includegraphics[width=15cm]{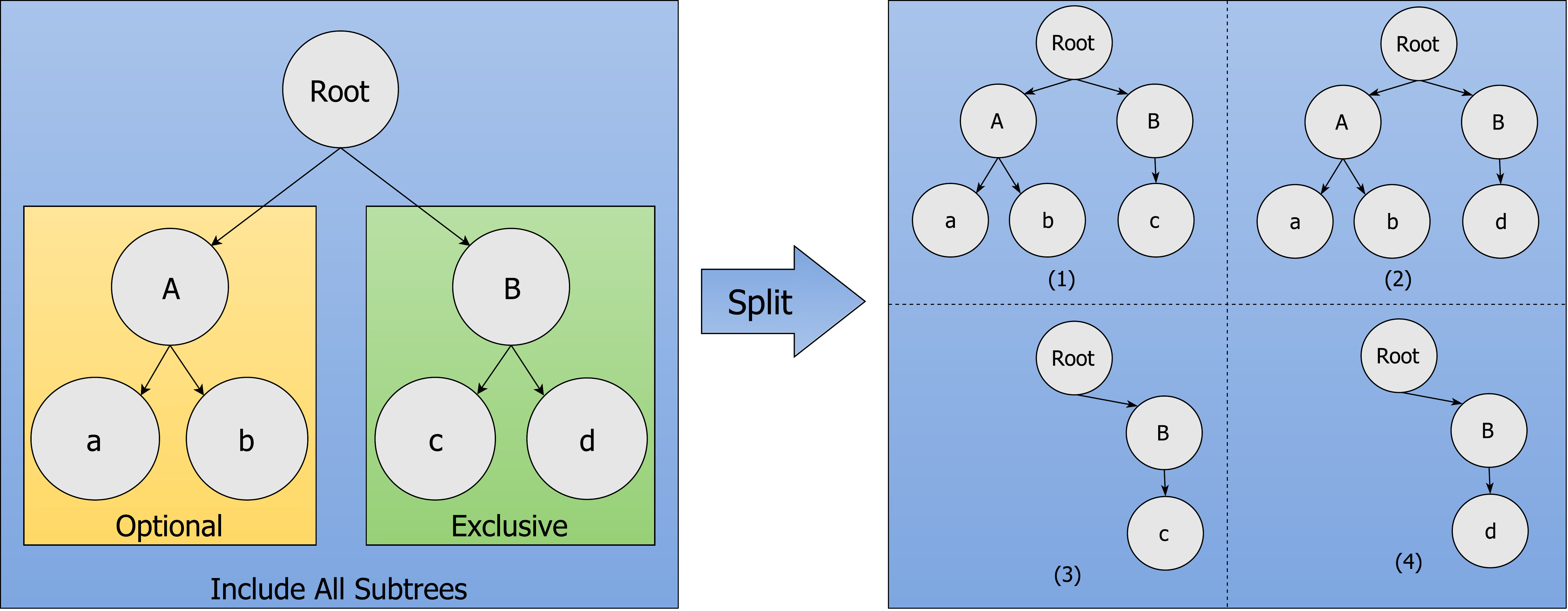}
    \caption{Splitting a specification tree into a set of disjoint trees. (Left) The specification tree contains two subtrees: an optional subtree $A$ and an exclusive subtree $B$. The root level indicates that some subset of both subtrees should be included if possible. (Right) The trees created by splitting the specification tree on the left. Because $A$ is flagged as optional, trees (1) and (2) include both children of $A$, while $A$ is excluded entirely from trees (3) and (4). Additionally, each of the four trees contains only one child of $B$. In terms of hyperparameter search, $A$ corresponds to an operation that is global across models, such as a preprocessing step, and $B$ corresponds to the disjoint models being searched.}
    \vspace{-1em}
    \label{fig:split}
\end{figure*}

Complexity offers a static measure of the need to search a particular space, gearing a search toward models with a large number of hyperparameters defined over a wide variety of spaces. Note that complexity is not necessarily an approximation of running time, rather it is a description of the size of the search spaces. In many cases, particularly when searching neural network architectures, there is a correspondence between hyperparameter count and running time, however this relationship does not generally hold.

\subsection{Priority}

To complement the static complexity heuristic, dynamic assessment of the need to search a model is also necessary. The priority heuristic accounts for model performance across different parametrizations, estimating the fitness of a model to learn from the data as a function of variation in performance.

Priority is calculated using the method described by Bergstra and Bengio~\cite{bergstra-2012-random-search} for approximating hyperparameter importance to a model's performance on a dataset. For a given model, a Gaussian process with RBF kernel is fit to a dataset consisting of the tested hyperparameter values and their resulting loss values. The learned length scale $l$ of the RBF kernel that maximizes the log marginal likelihood is extracted after fitting as an approximation of the sensitivity of the RBF kernel to changes in the hyperparameter values. This process is repeated 50 times to obtain a sample of length scales, $L$, and the priority is approximated as
\begin{equation} \label{eqn:priority}
P(L) = \min(L)^{-1} - \max(L)^{-1}
\end{equation}

In essence, $P(L)$ approximates the covariance between parametrizations and their resulting performance to give an estimate of the intrinsic fitness of a model for the learning task. Models with high parameterization covariance (low $P(L)$) are less pressing to search because their fitness is known; those with low covariance (high $P(L)$) have a wide range of observed performance and thus should be searched more thoroughly to determine their fitness. Note that a model with ``consistent performance" will receive a low priority regardless of whether it performs well or poorly, as priority is a measure of variation.

Like complexity, models with a larger priority are preferred in the search because there is less certainty about their fitness to the learning task. This lack of certainty indicates that the hyperparameter spaces should be explored more thoroughly to better understand how the model performs under different parameterizations. Models with low priority should continue to be searched, given that the performance of the model over the entire hyperparameter domain cannot be known, however the number of searches allocated should be scaled back.

\section{SHADHO Framework}

To create a heuristic scheduler for hyperparameter optimization, we implemented complexity and priority in the SHADHO framework. SHADHO is an open-source Python package built on top of the scientific Python stack~\cite{jones-2001-scipy,pedregosa-2011-scikit-learn} and the Work Queue framework~\cite{yi-2009-work-queue} for hyperparameter generation and distributed task management, respectively.

\subsection{Defining Search Spaces}

Prior to a search, models are defined as a tree with hyperparameter domains as the leaves. In this tree, individual models are demarcated by tagging subtrees as ``exclusive" (only one path below may be followed at a time) or ``optional" (the subtree is either included or excluded entirely). SHADHO splits this tree into a forest at runtime based on the exclusive and optional tags, with each tree in the forest corresponding to a single model, as shown in Figure~\ref{fig:split}.

Search space definition semantics were created based on the best practices outlined by Bergstra \textit{et al.}~\cite{bergstra-2015-hyperopt}. As a result, the API will be familiar to users of Hyperopt, a standard open-source distributed hyperparameter optimization framework. Examples of defining search spaces, including the usage of ``exclusive" and ``optional" flags, can be found in the SHADHO documentation (see supp. material).

\subsection{Hardware Awareness with Compute Classes}

To simplify hardware-aware scheduling, SHADHO groups connected workers based on common hardware resources. These \textit{compute classes} can be grouped by GPU model, number of cores, memory size, and other arbitrary user-defined features. These are then ranked in order of performance to create a hierarchy by which models may be assigned to hardware.

\subsection{Using Heuristics in SHADHO}

SHADHO uses an implementation of the complexity and priority heuristics defined in Equations~\ref{eqn:complexity} and~\ref{eqn:priority} to rank each tree (model) in the forest. The models are then ranked by their complexity and priority, with equal precedence given to both heuristics. The ranking is used to weight models such that higher-ranked models are scheduled more often and assigned to higher-performing hardware. Considering both heuristics as equal balances the search between emphasizing models with more potential parametrizations and those with greater performance variation across parametrizations.

While SHADHO uses both heuristics by default, one or both may be deactivated to accommodate the model search and available resources. For example, if a large model cannot feasibly run on one particular compute class, the search can use complexity only to ensure that the model is only trained and evaluated on nodes of that compute class. Similarly, exploratory studies with a large number of models may benefit from using priority only to direct the search based on model performance alone. For evaluation purposes, SHADHO may also operate in heuristic-free mode (SHADHO-FCFS), in which models are scheduled FCFS. SHADHO-FCFS is equivalent to other distributed hyperparameter optimization frameworks in that it makes no decisions about task scheduling.

\begin{figure}[!t]
\centering
\includegraphics[width=\columnwidth]{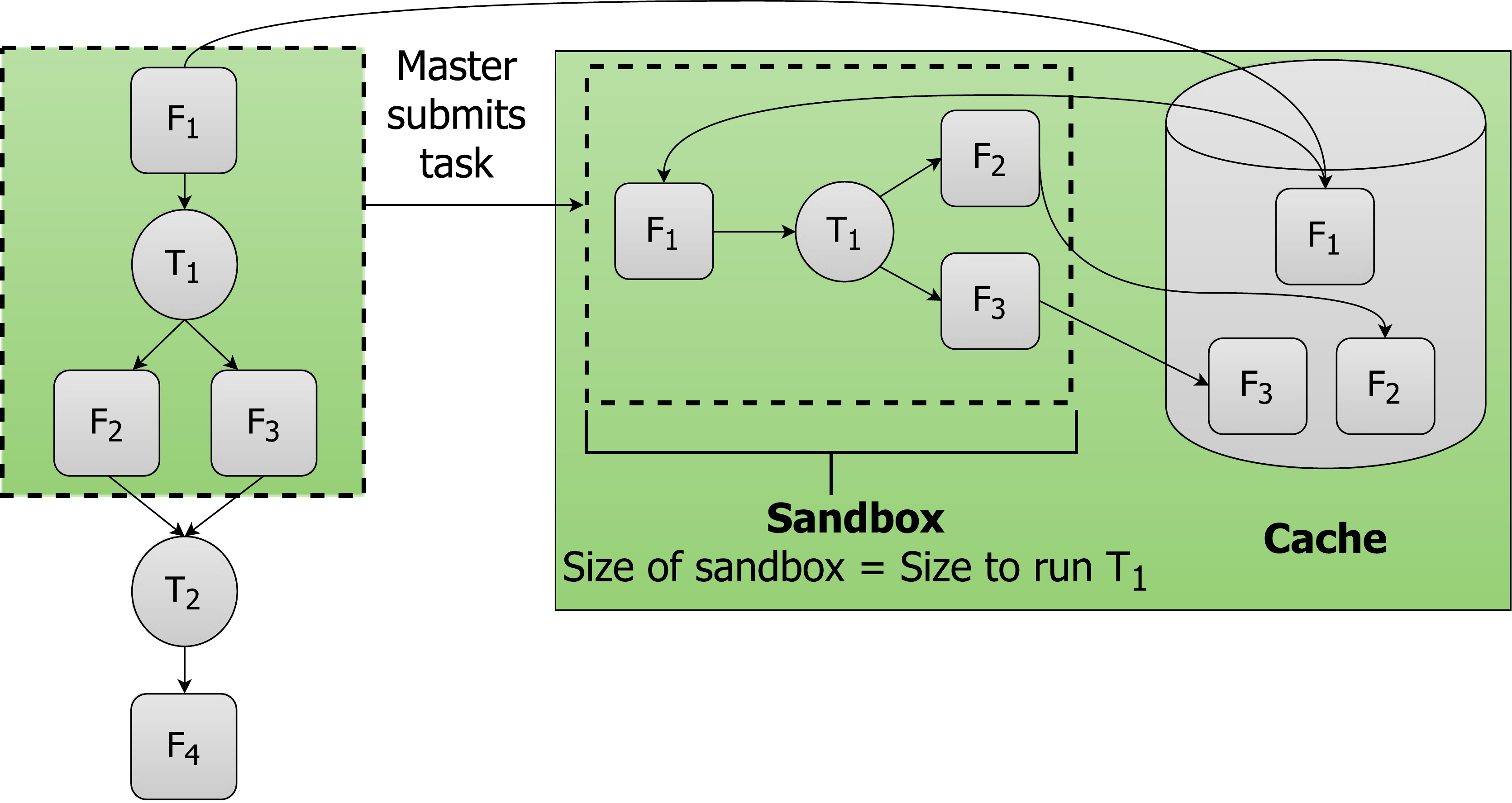}
\caption{Work Queue Master-Worker architecture. Tasks are dispatched by the master process with the necessary input files ($F_{1}$), command to run ($T_{1}$), and the expected output files ($F_{2}$ and $F_{3}$) to an available worker process. Each worker may handle multiple tasks that all share a common data cache in each worker.}
\vspace{-1em}
\label{fig:master_worker}
\end{figure}

\subsection{Distributed Task Management}

Distributed task management and execution in SHADHO are handled by the Work Queue execution engine~\cite{yi-2009-work-queue}. Work Queue uses a master-worker scheme consisting of a centralized master and distributed worker processes. The master coordinates the workers that may be running on a variety of machines in clusters, clouds, or grids. Work Queue workers are %pilot jobs that are
persistent processes submitted to a batch job system that communicate with the master to request work. %Once the worker is scheduled by the batch system, it will attempt to communicate with the master, asking for work.
If tasks are available, the master dispatches them as shown in Figure~\ref{fig:master_worker}. The worker operates entirely within a sandboxed environment on its host system that includes cached versions of files used across tasks. % The sandbox contains all files specified for the task, and the worker will cache files between tasks if requested. % If a worker does not receive new work within a certain timeframe it will clean up the sandbox and cache, then shut down.

%Work Queue has several performance optimizations that other hyperparameter optimization applications may not be able to support. First, each worker has a local cache which persists for the worker's lifetime. The cache prevents common input files from being transferred excessively, which in turn will reduce the time it takes for a task to complete. Work Queue also tries to match tasks to workers that already have some or all relevant input files present in their respective caches.

Workers and tasks support specifying hardware resource requirements. For the worker process, the resource requirements may consist of any or all of: cores, memory, disk, and GPUs. This ensures that the batch system managing the workers  will only land each worker process on a machine with at least the minimum requested resources. A worker can also advertise an arbitrary feature to the master (i.e., the CPU model). For tasks, the resource specifications ensure the master only dispatches tasks to workers that can handle them. Task resource specification includes the same resources as the worker (cores, memory, disk, and GPUs), but it may also request any arbitrary feature. %By requesting an arbitrary feature, we provide a soft guarantee that the master will dispatch tasks to workers with the matching feature.
SHADHO uses Work Queue's hardware requirements specification to group workers into compute classes.

\section{Experiments}

To demonstrate the effectiveness of heuristic-based hardware assignment for hyperparameter search, we apply SHADHO to two computer vision problems: Support Vector Machine (SVM) optimization for handwritten character identification, and fully-convolutional neural network (FCNN) optimization for cell segmentation in electron microscopy images. SVM optimization allows us to quantify improvements in performance compared to software with FCFS scheduling. FCNN optimization demonstrates SHADHO's applicability to difficult learning problems.

\subsection{SVM Optimization}

\begin{figure}
    \centering
    \includegraphics[width=\columnwidth]{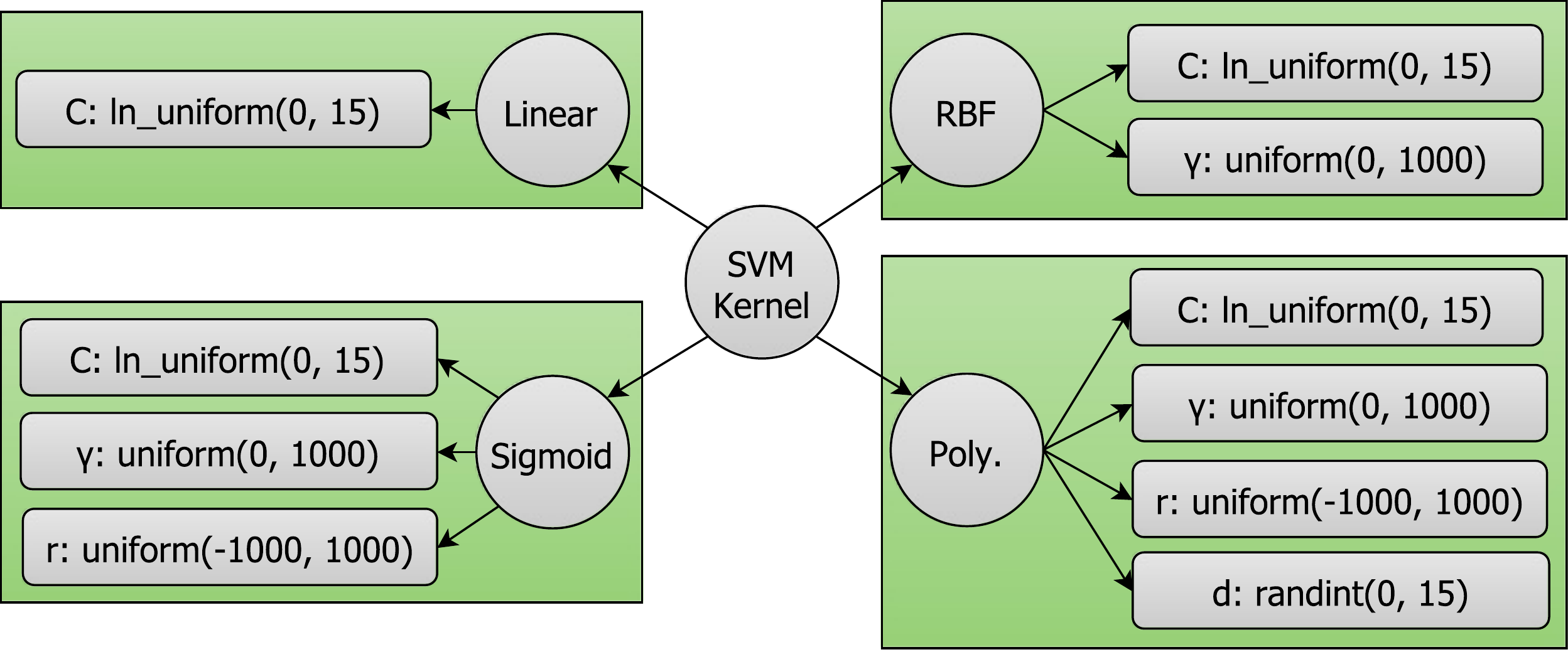}
    \caption{Specification for searching four SVM kernel functions, with four distinct models. Each kernel contains a different number of hyperparameters, giving the trees a strong complexity-based ordering. $C$: soft-margin constant, defined uniformly over $[0, 15]$, and scaled logarithmically. $\gamma$: kernel coefficient, defined uniformly over $(0, 10^{3}]$. $r$: an additional coefficient, defined over $[-10^{3}, 10^{3}]$. $d$: polynomial degree, a random integer from 1 -- 15.}
    \vspace{-1em}
    \label{fig:svmspec}
\end{figure}

Our first experiment, optimizing SVM for handwritten digit recognition using the MNIST dataset \cite{lecun-1998-mnist}, is a benchmark for determining the throughput of hyperparameter optimization software. While MNIST classification is a solved problem, it is a useful benchmark for demonstrating increases to throughput enabled by SHADHO because each individual sets of hyperparameters may be tested quickly, but performance and running times vary across SVM kernels. Figure~\ref{fig:svmspec} presents the four SVM kernels and the spaces searched for each of their hyperparameters.

\begin{figure*}
    \centering
    \mbox{
        \includegraphics[width=\columnwidth]{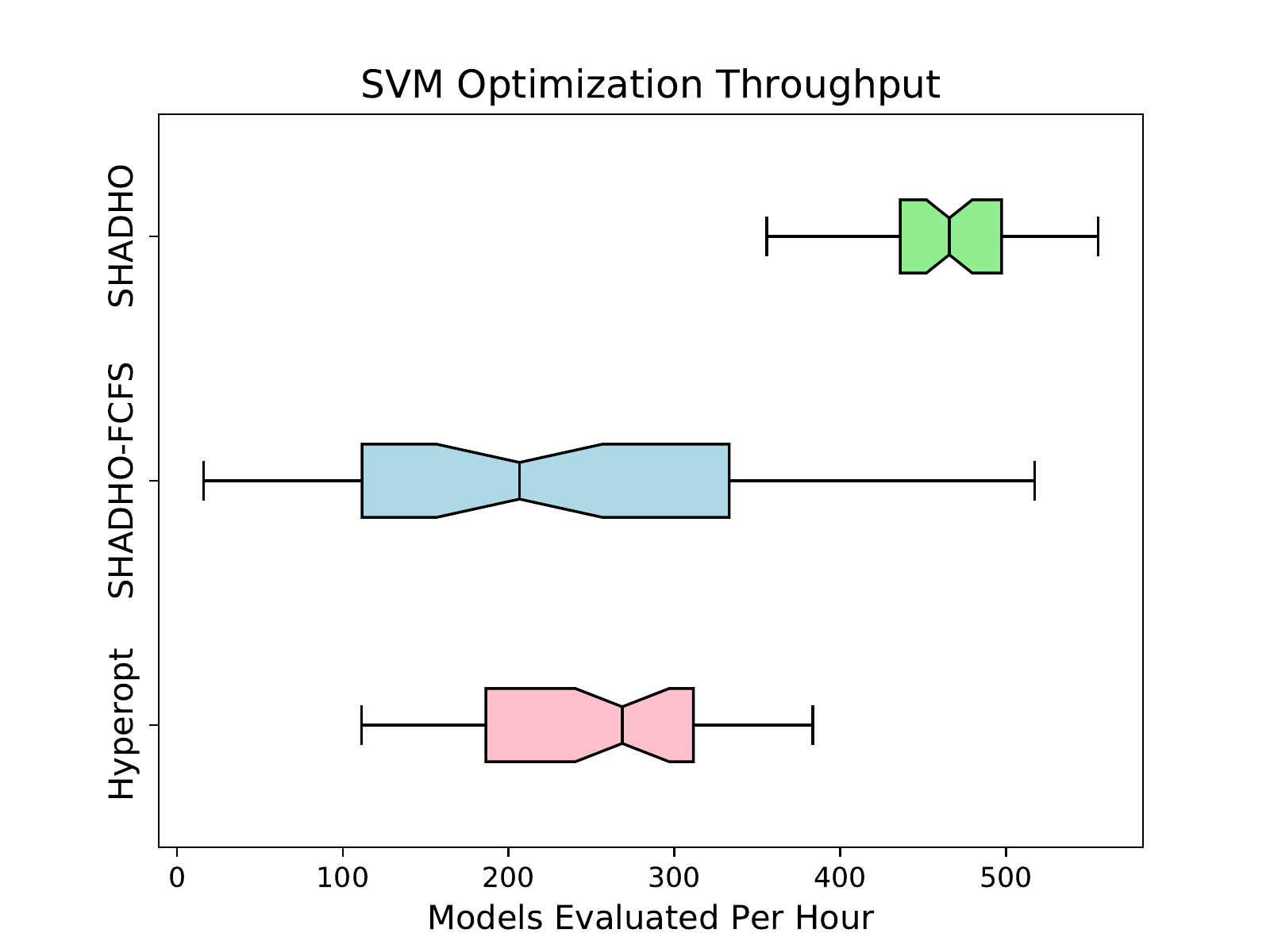}
        \includegraphics[width=\columnwidth]{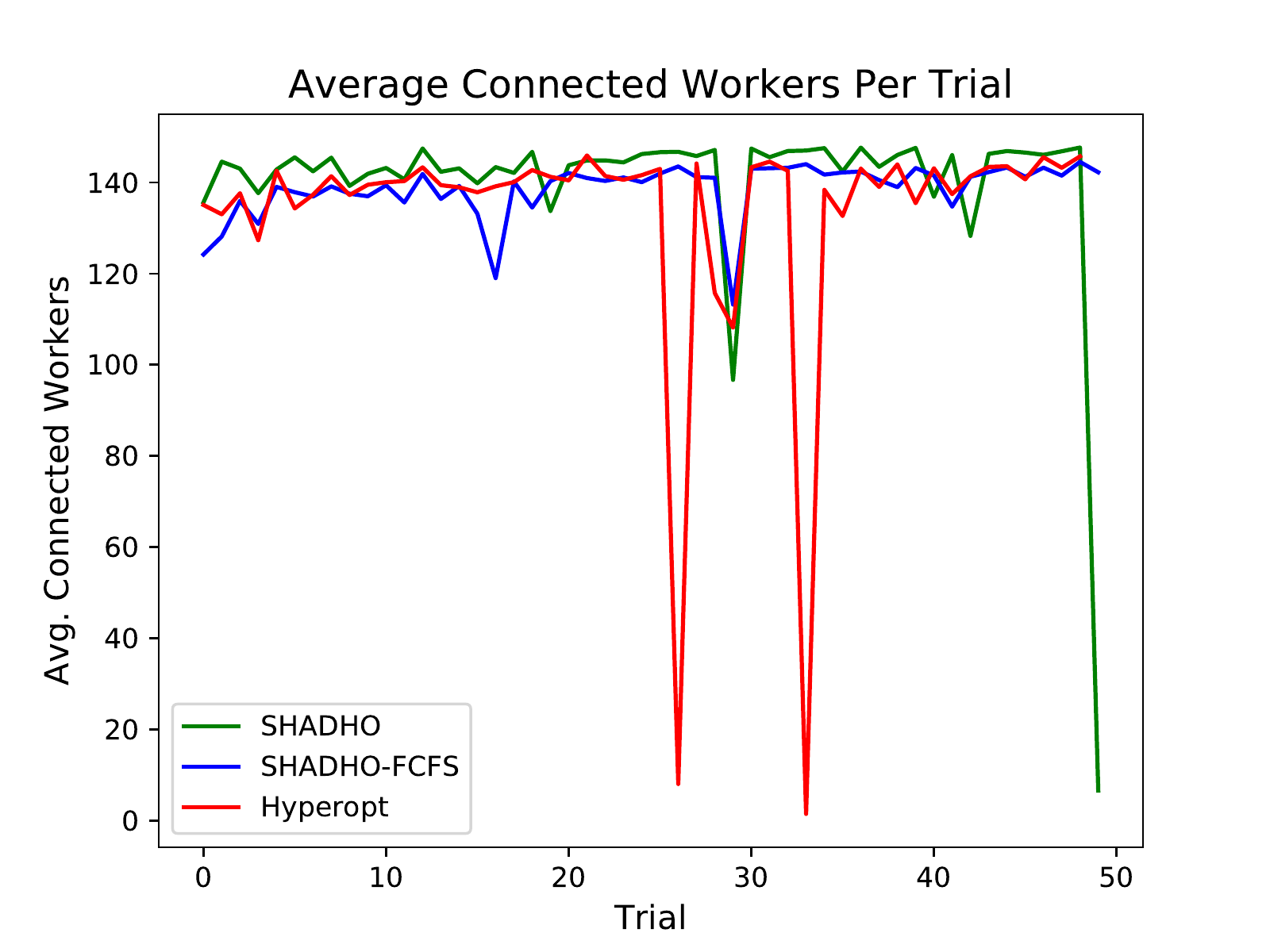}
    }
    \caption{The throughput in tasks completed per hour over all SVM optimization trials (left) and average connected workers over the first hour of each trial (right) for SHADHO (green), SHADHO-FCFS (blue), and Hyperopt~\cite{bergstra-2013-hyperopt} (red). SHADHO had the highest throughput on average, $1.8\times$ that of Hyperopt and $2.0\times$ that of SHADHO-FCFS. SHADHO also had the smallest throughput distribution, indicating that heuristic-based distribution leads to more stable performance. In most cases, the difference in average number of connected workers between SHADHO, SHADHO-FCFS and Hyperopt in a given trial is extremely small, indicating that the three methods were operating with approximately the same set of distributed workers in every case. Thus, the difference in throughput is attributable to heuristic-based task scheduling except in extreme cases of network load (e.g., trials 26, 29, 33, and 48).}
    \vspace{-1em}
    \label{fig:svmresults}
\end{figure*}

In this experiment, we used the \texttt{scikit-learn} \cite{pedregosa-2011-scikit-learn} SVM implementation trained using one-vs-rest classification with one estimator per class (10 total) parallelized over the number of available cores on a worker. Tests were distributed over 150 workers: 50 4-core machines, 50 8-core machines, and 50 16-core machines. In this case, a 4-core worker was able to train 4 estimators in parallel, while a 16-core worker trained all 10 simultaneously.

We performed random search using SHADHO, SHADHO-FCFS, and Hyperopt~\cite{bergstra-2013-hyperopt}, a standard hyperparameter optimization package, and report the performance in terms of the number of hyperparameters tested per hour. To standardize each trial, SHADHO performed a search for one hour, assigning searches to workers based on their complexity and priority. The same set of parameters were then tested using SHADHO-FCFS and Hyperopt, both of which perform FCFS scheduling.

\begin{table}[h]
    \centering
    \caption{SVM Optimization Metrics}
    \begin{tabular}{|c|c|}
        \hline
        \textbf{Method} & \textbf{Avg. Throughput (tasks/hr)} \\ \hline
        SHADHO & 463.03 \\ \hline
        SHADHO-FCFS & 227.28 \\ \hline
        Hyperopt~\cite{bergstra-2013-hyperopt} & 252.42 \\ \hline
    \end{tabular}
    \label{tab:svm_metrics}
\end{table}

The average throughput per trial over 48 trials is shown in Table~\ref{tab:svm_metrics}. On average, SHADHO achieved a $1.8\times$ increase in throughput over Hyperopt and a $2\times$ increase over SHADHO-FCFS, with all three using approximately the same workers in each trial. As shown in Figure~\ref{fig:svmresults}, the distribution of SHADHO's throughput across the 48 trials was much smaller than that of Hyperopt or SHADHO-FCFS, indicating that heuristic-based hardware-aware task scheduling leads to more consistent performance than FCFS scheduling. We attempted to mitigate the effects of distributing trials across a network with dynamic usage patterns by running the three methods back-to-back in each trial. In general, this led to similar worker connectivity patterns, however sharp drops in connectivity can be seen in Figure~\ref{fig:svmresults}. In each of these cases, the limiting factor on throughput was network load --- we intentionally ran these experiments in a production environment with other users.

It should be noted that both SHADHO-FCFS and Hyperopt were able to match the throughput of SHADHO during a small number of trials. This is not unexpected: a FCFS scheduling scheme, which is effectively random task scheduling, will on occasion achieve much higher throughput than average due to task order. This is not reliable, though, as is evident from the size of each distribution. Both of the FCFS-based methods have wide throughput distributions, indicating extremely unreliable performance. SHADHO, on the other hand, boasts a smaller throughput distribution and thus more consistent performance.

%We tested SHADHO against Hyperopt and SHADHO-FCFS to show that the increase in throughput is a result of heuristic task assignment rather than a change in backend technology. SHADHO-FCFS and Hyperopt use the same scheduling scheme, assigning a task to the first available worker. Heuristic assignment by SHADHO results in double the throughput of both of these naive scheduling policies. Thus, we conclude that the use of complexity- and priority-based ranking for scheduling is well-suited to increase the performance of model selection and hyperparameter optimization tasks.

\subsection{U-Net Optimization}

\begin{figure*}[!tp]
\centering
    \includegraphics[width=\textwidth]{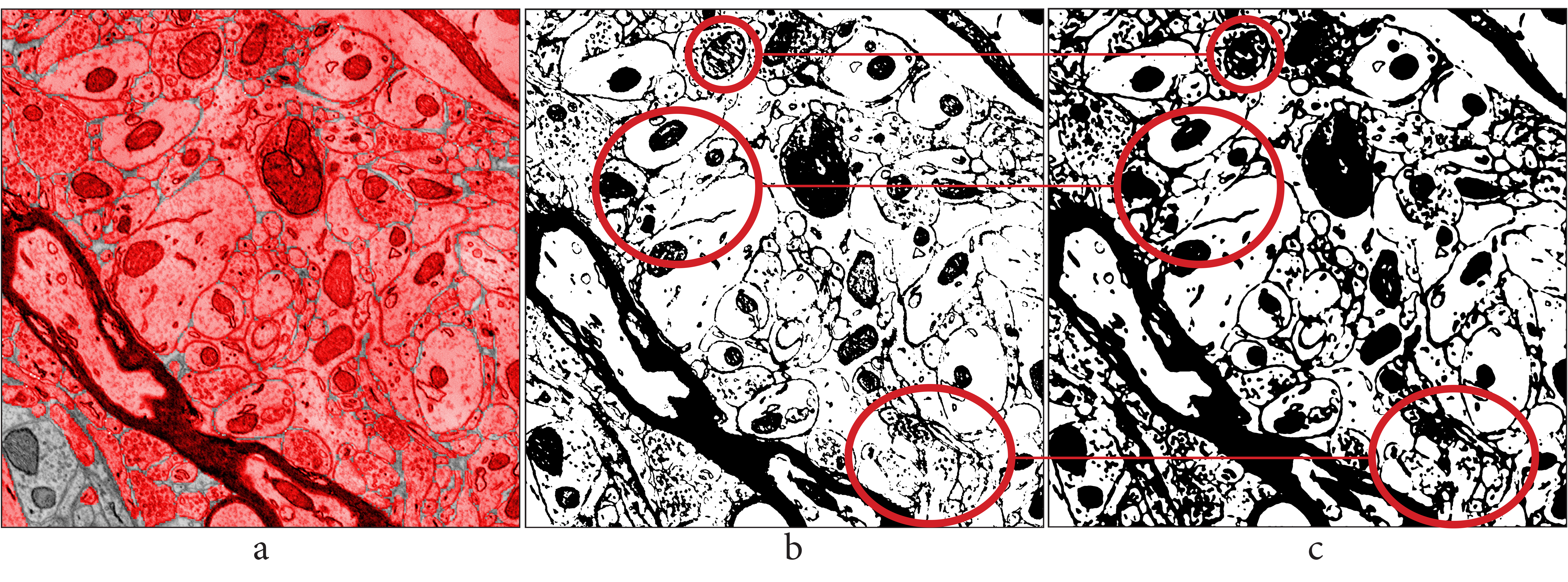}
    \caption{An image from the validation set created from the EM cell segmentation dataset of Kasthuri \textit{et al.}~\cite{kasthuri-2015-cell} with ground-truth overlaid in red (a), and predictions obtained from the standard U-Net model (b) and an optimized model found by SHADHO (c). The SHADHO model includes less noise in its predictions and more clearly separates the individual cells. Both segmentations classify dark features in the cell interiors as background, and this is a difficult challenge to overcome due to their similarity to the background of the ground-truth. Circled: Examples of merge errors and misclassifications improved upon by our discovered model. Best viewed in color.}
    \vspace{-1em}
    \label{fig:unet_result}
\end{figure*}

In the neuroscientific domain, neuronal cell segmentation by membrane detection in electron microscopic (EM) images is a difficult problem. In general, EM data is noisy with staining and imaging artifacts obscuring features, and few annotated datasets exist to train a pixel-wise classifier~\cite{helmstaedter-2013-connectomics-challenges,lichtman-2014-big-data-challenges}. We selected the EM dataset introduced by Kasthuri \textit{et al.}~\cite{kasthuri-2015-cell} because it is one of the few with comprehensive ground-truth annotations created by a domain expert that is not of trivial size. We selected a $253 \times 2048 \times 2048$ voxel subset of the data for training and evaluation. The goal of this experiment was to train a pixel-wise classifier to correctly segment cells in the volume.

We used SHADHO to optimize U-Net~\cite{ronneberger-2015-unet}, a state-of-the-art fully-convolutional neural network for microscopic image segmentation. In addition to a forward downsampling / upsampling path, U-Net includes bypass connections that feed the output of early convolutional layers on the downsampling path to the corresponding convolutional unit on the upsampling path. In this experiment, we conducted model search and hyperparameter optimization over a number of different U-Net models with varying bypass connections. Table~\ref{tbl:unet_params} lists the hyperparameter values searched, with all hyperparameters except learning rate randomly sampled for each convolutional layer. In all cases, models were trained to optimize cross entropy on augmented tiles from the first 80$\%$ of the dataset and evaluated on the last 20$\%$ of the dataset. The ground truth was inverted to emphasize cell membranes over cell interiors. Models were trained for 150 epochs with a batch size of 100 $128 \times 128$-pixel images and early stopping for models with plateauing performance.

\begin{table}[h]
\centering
\caption{U-Net Hyperparameter Search Spaces}
\begin{tabular}{ | c | p{5cm} | }
 \hline
 \textbf{Parameter} & \textbf{Values} \\ \hline
 Min. Kernels & 16, 32, 64, 128 \\  \hline
 Kernel Size &  1, 3, 5, 7, 9 \\  \hline
 Activations & sigmoid, tanh, relu, elu, PReLU, LeakyReLU, \newline ThresholdedReLU \\  \hline
 Initializers & zeros, ones, glorot\_normal, he\_normal \\ \hline
 Regularizers & l1, l2, l1\_l2 \\  \hline
 Dropout Rate & uniform distribution over $[0, 1]$ \\  \hline
 Learning Rate & uniform distribution over $[10^{-4}, 1]$ \\
 \hline
\end{tabular}
\label{tbl:unet_params}
\vspace{-1em}
\end{table}

Over the course of one week, we evaluated 1075 U-Net models with different parametrizations using 16 NVIDIA Titan X Pascal, 50 NVIDIA Tesla K80, and 8 NVIDIA GTX 1080ti GPUs\footnote{Additional details about this experiment may be found in supp. mat.}. We compared these against a standard U-Net model as described by Ronneberger \textit{et al.}~\cite{ronneberger-2015-unet} trained using a cross entropy loss function and the Adam optimizer with a learning rate of $10^{-4}$. The published U-Net achieved a validation loss of 1.07 after training, and we discovered 515 models with a lower loss. An example of the difference in classification ability between one of our models and the published U-Net model is shown in Figure~\ref{fig:unet_result}. As can be seen, the U-Net model discovered by SHADHO increases the separability between cells in the image and incurred fewer merge errors and misclassifications than the standard U-Net model. Based on these results, we conclude that SHADHO was able to find higher-quality models than the hand-tuned published U-Net model.

\section{Conclusions}

Hyperparameter optimization is a crucial step in the machine learning process, and evaluating as many parametrizations as possible increases the chance of finding a high-quality model. SHADHO increases the throughput of hyperparameter optimization by determining the complexity and priority of searching each model and adjusting the proportion of searches allocated to each by assigning high-priority, high-complexity models to more performant hardware. In the case of SVM kernel optimization, SHADHO increased the throughput of model evaluations by a factor of 2.0 over distributed hyperparameter optimization without hardware awareness. Moreover, when applied to a U-Net for microscopic image segmentation, SHADHO discovered a large number of parametrizations with lower validation error than the published U-Net, indicating that SHADHO is able to improve upon hand-tuned models.

The current implementation of SHADHO only makes use of a random hyperparameter search strategy. Recent studies have introduced promising search strategies involving Bayesian optimization and bandit-based search, and future versions of SHADHO will incorporate these and forthcoming search strategy developments. Additionally, the problem of automated neural network architecture construction has been broached in recent deep learning work~\cite{valdez-2014-modular,zoph-2016-archsearch}. One of the prime benefits of SHADHO is the ability to dynamically reallocate work to appropriate hardware. Neural network architecture search involves testing neural networks of varying complexities and running times, making SHADHO a suitable framework for exploring and scaling their architectures.

The ability to effectively allocate hyperparameters to hardware is central to SHADHO's operation. We will continue to explore other methods for approximating complexity and priority, including methods for understanding running time across heterogeneous hardware (complexity) and comparative performance metrics between models and search spaces (priority). SHADHO will allow us to explore these methods at a massive scale and advance both machine learning and distributed computing.

% The submission is blind
\section{Acknowledgements}

This research was supported in part by the Notre Dame Center for Research Computing through access to distributed computing resources. This research used resources of the Argonne Leadership Computing Facility, which is a DOE Office of Science User Facility supported under Contract DE-AC02-06CH11357. Funding was provided under IARPA contract \#D16PC00002.

{\small
\bibliographystyle{ieee}
\bibliography{shadho}

\begin{thebibliography}{10}\itemsep=-1pt

\bibitem{alebrahim-2017-task-scheduling}
S.~{Al Ebrahim} and I.~Ahmad.
\newblock Task scheduling for heterogeneous computing systems.
\newblock {\em The Journal of Supercomputing}, 73(6):2313--2338, 2017.

\bibitem{benardo-2007-optimizing}
P.~Benardos and G.-C. Vosniakos.
\newblock Optimizing feedforward artificial neural network architecture.
\newblock {\em Engineering Applications of Artificial Intelligence},
  20(3):365--382, 2007.

\bibitem{bergstra-2012-random-search}
J.~Bergstra and Y.~Bengio.
\newblock Random search for hyper-parameter optimization.
\newblock {\em Journal of Machine Learning Research}, 13(Feb):281--305, 2012.

\bibitem{bergstra-2015-hyperopt}
J.~Bergstra, B.~Komer, C.~Eliasmith, D.~Yamins, and D.~D. Cox.
\newblock Hyperopt: a python library for model selection and hyperparameter
  optimization.
\newblock {\em Computational Science \& Discovery}, 8(1):014008, 2015.

\bibitem{bergstra-2013-hyperopt}
J.~Bergstra, D.~Yamins, and D.~D. Cox.
\newblock Hyperopt: A python library for optimizing the hyperparameters of
  machine learning algorithms.
\newblock In {\em Proceedings of the 12th Python in Science Conference}, pages
  13--20. Citeseer, 2013.

\bibitem{bergstra-2011-tpe}
J.~S. Bergstra, R.~Bardenet, Y.~Bengio, and B.~K{\'e}gl.
\newblock Algorithms for hyper-parameter optimization.
\newblock In {\em Advances in Neural Information Processing Systems}, pages
  2546--2554, 2011.

\bibitem{biswas-2017-multiqueue}
T.~Biswas, P.~Kuila, and A.~K. Ray.
\newblock Multi-level queue for task scheduling in heterogeneous distributed
  computing system.
\newblock In {\em Advanced Computing and Communication Systems (ICACCS), 2017
  4th International Conference on}, pages 1--6. IEEE, 2017.

\bibitem{heft-lookahead-2010}
L.~F. Bittencourt, R.~Sakellariou, and E.~R.~M. Madeira.
\newblock Dag scheduling using a lookahead variant of the heterogeneous
  earliest finish time algorithm.
\newblock In {\em 2010 18th Euromicro Conference on Parallel, Distributed and
  Network-based Processing}, pages 27--34, Feb 2010.

\bibitem{DBLP:journals/corr/BojarskiTDFFGJM16}
M.~Bojarski, D.~D. Testa, D.~Dworakowski, B.~Firner, B.~Flepp, P.~Goyal, L.~D.
  Jackel, M.~Monfort, U.~Muller, J.~Zhang, X.~Zhang, J.~Zhao, and K.~Zieba.
\newblock End to end learning for self-driving cars.
\newblock {\em CoRR}, abs/1604.07316, 2016.

\bibitem{brock2017smash}
A.~Brock, T.~Lim, J.~M. Ritchie, and N.~Weston.
\newblock Smash: one-shot model architecture search through hypernetworks.
\newblock {\em arXiv preprint arXiv:1708.05344}, 2017.

\bibitem{claesen-2014-optunity}
M.~Claesen, J.~Simm, D.~Popovic, and B.~Moor.
\newblock Hyperparameter tuning in {P}ython using optunity.
\newblock In {\em Proceedings of the International Workshop on Technical
  Computing for Machine Learning and Mathematical Engineering}, 2014.

\bibitem{cortes-2016-adanet}
C.~Cortes, X.~Gonzalvo, V.~Kuznetsov, M.~Mohri, and S.~Yang.
\newblock Adanet: Adaptive structural learning of artificial neural networks.
\newblock {\em arXiv preprint arXiv:1607.01097}, 2016.

\bibitem{dewancker-2016-sigopt}
I.~Dewancker, M.~McCourt, S.~Clark, P.~Hayes, A.~Johnson, and G.~Ke.
\newblock Evaluation system for a bayesian optimization service.
\newblock {\em arXiv preprint arXiv:1605.06170}, 2016.

\bibitem{domhan-2015-learning-curve}
T.~Domhan, J.~T. Springenberg, and F.~Hutter.
\newblock Speeding up automatic hyperparameter optimization of deep neural
  networks by extrapolation of learning curves.
\newblock In {\em IJCAI}, pages 3460--3468, 2015.

\bibitem{duan-2005-grid}
K.-B. Duan and S.~S. Keerthi.
\newblock Which is the best multiclass svm method? an empirical study.
\newblock In {\em International Workshop on Multiple Classifier Systems}, pages
  278--285. Springer, 2005.

\bibitem{esteva2017dermatologist}
A.~Esteva, B.~Kuprel, R.~A. Novoa, J.~Ko, S.~M. Swetter, H.~M. Blau, and
  S.~Thrun.
\newblock Dermatologist-level classification of skin cancer with deep neural
  networks.
\newblock {\em Nature}, 542(7639):115--118, 2017.

\bibitem{fergus2013}
R.~Fergus.
\newblock Deep learning for computer vision, 2013.
\newblock Tutorial Presented at NIPS 2013.

\bibitem{fernandes-2005-trial-and-error}
F.~Fernandes and L.~Lona.
\newblock Neural network applications in polymerization processes.
\newblock {\em Brazilian Journal of Chemical Engineering}, 22(3):401--418,
  2005.

\bibitem{friedrichs-2005-evolutionary}
F.~Friedrichs and C.~Igel.
\newblock Evolutionary tuning of multiple svm parameters.
\newblock {\em Neurocomputing}, 64:107--117, 2005.

\bibitem{furtuna-2011-optimization}
R.~Furtuna, S.~Curteanu, and M.~Cazacu.
\newblock Optimization methodology applied to feed-forward artificial neural
  network parameters.
\newblock {\em International Journal of Quantum Chemistry}, 111(3):539--553,
  2011.

\bibitem{garro-2015-particle-swarm}
B.~A. Garro and R.~A. V{\'a}zquez.
\newblock Designing artificial neural networks using particle swarm
  optimization algorithms.
\newblock {\em Computational intelligence and neuroscience}, 2015:61, 2015.

\bibitem{helmstaedter-2013-connectomics-challenges}
M.~Helmstaedter.
\newblock Cellular-resolution connectomics: challenges of dense neural circuit
  reconstruction.
\newblock {\em Nature methods}, 10(6):501--507, 2013.

\bibitem{hu-2016-pruning}
H.~Hu, R.~Peng, Y.-W. Tai, and C.-K. Tang.
\newblock Network trimming: A data-driven neuron pruning approach towards
  efficient deep architectures.
\newblock {\em arXiv preprint arXiv:1607.03250}, 2016.

\bibitem{hutter-2013-smac}
F.~Hutter, H.~Hoos, and K.~Leyton-Brown.
\newblock An evaluation of sequential model-based optimization for expensive
  blackbox functions.
\newblock In {\em Proceedings of the 15th annual conference companion on
  Genetic and evolutionary computation}, pages 1209--1216. ACM, 2013.

\bibitem{ilievski-2017-rbf}
I.~Ilievski, T.~Akhtar, J.~Feng, and C.~A. Shoemaker.
\newblock Efficient hyperparameter optimization for deep learning algorithms
  using deterministic rbf surrogates.
\newblock In {\em AAAI}, pages 822--829, 2017.

\bibitem{jones-1998-smbo}
D.~R. Jones, M.~Schonlau, and W.~J. Welch.
\newblock Efficient global optimization of expensive black-box functions.
\newblock {\em Journal of Global optimization}, 13(4):455--492, 1998.

\bibitem{jones-2001-scipy}
E.~Jones, T.~Oliphant, P.~Peterson, et~al.
\newblock {SciPy}: Open source scientific tools for {Python}, 2001--.
\newblock [Online; accessed 2017-02-23].

\bibitem{kasthuri-2015-cell}
N.~Kasthuri, K.~J. Hayworth, D.~R. Berger, R.~L. Schalek, J.~A. Conchello,
  S.~Knowles-Barley, D.~Lee, A.~V{\'a}zquez-Reina, V.~Kaynig, T.~R. Jones,
  et~al.
\newblock Saturated reconstruction of a volume of neocortex.
\newblock {\em Cell}, 162(3):648--661, 2015.

\bibitem{kotthoff-2016-autoweka}
L.~Kotthoff, C.~Thornton, H.~H. Hoos, F.~Hutter, and K.~Leyton-Brown.
\newblock Auto-weka 2.0: Automatic model selection and hyperparameter
  optimization in weka.
\newblock {\em Journal of Machine Learning Research}, 17:1--5, 2016.

\bibitem{lecun-1998-mnist}
Y.~LeCun, C.~Cortes, and C.~J. Burges.
\newblock The mnist database of handwritten digits, 1998.

\bibitem{li-2016-bandit}
L.~Li, K.~Jamieson, G.~DeSalvo, A.~Rostamizadeh, and A.~Talwalkar.
\newblock Efficient hyperparameter optimization and infinitely many armed
  bandits.
\newblock In {\em ICML 2016 workshop on AutoML (AutoML 2016)}, 2016.

\bibitem{lichtman-2014-big-data-challenges}
J.~W. Lichtman, H.~Pfister, and N.~Shavit.
\newblock The big data challenges of connectomics.
\newblock {\em Nature neuroscience}, 17(11):1448--1454, 2014.

\bibitem{maclaurin-2015-gradient}
D.~Maclaurin, D.~K. Duvenaud, and R.~P. Adams.
\newblock Gradient-based hyperparameter optimization through reversible
  learning.
\newblock In {\em ICML}, pages 2113--2122, 2015.

\bibitem{Parkhi15}
O.~M. Parkhi, A.~Vedaldi, and A.~Zisserman.
\newblock Deep face recognition.
\newblock In {\em British Machine Vision Conference}, 2015.

\bibitem{pedregosa-2011-scikit-learn}
F.~Pedregosa, G.~Varoquaux, A.~Gramfort, V.~Michel, B.~Thirion, O.~Grisel,
  M.~Blondel, P.~Prettenhofer, R.~Weiss, V.~Dubourg, J.~Vanderplas, A.~Passos,
  D.~Cournapeau, M.~Brucher, M.~Perrot, and E.~Duchesnay.
\newblock Scikit-learn: Machine learning in {P}ython.
\newblock {\em Journal of Machine Learning Research}, 12:2825--2830, 2011.

\bibitem{ronneberger-2015-unet}
O.~Ronneberger, P.~Fischer, and T.~Brox.
\newblock U-net: Convolutional networks for biomedical image segmentation.
\newblock In {\em International Conference on Medical Image Computing and
  Computer-Assisted Intervention}, pages 234--241. Springer, 2015.

\bibitem{Silva:2008:HRA:1462704.1462713}
J.~a.~N. Silva, L.~Veiga, and P.~Ferreira.
\newblock Heuristic for resources allocation on utility computing
  infrastructures.
\newblock In {\em Proceedings of the 6th International Workshop on Middleware
  for Grid Computing}, MGC '08, pages 9:1--9:6, New York, NY, USA, 2008. ACM.

\bibitem{snoek-2012-spearmint}
J.~Snoek, H.~Larochelle, and R.~P. Adams.
\newblock Practical bayesian optimization of machine learning algorithms.
\newblock In {\em Advances in neural information processing systems}, pages
  2951--2959, 2012.

\bibitem{aetros-2016}
A.~Team.
\newblock Aetros, 2016.

\bibitem{h2oai-2015}
T.~H. Team.
\newblock H2o: Scalable machine learning. version 3.1.0.99999, 2015.

\bibitem{heft-2002}
H.~Topcuoglu, S.~Hariri, and M.-Y. Wu.
\newblock Performance-effective and low-complexity task scheduling for
  heterogeneous computing.
\newblock {\em IEEE Transactions on Parallel and Distributed Systems},
  13(3):260--274, Mar 2002.

\bibitem{valdez-2014-modular}
F.~Valdez, P.~Melin, and O.~Castillo.
\newblock Modular neural networks architecture optimization with a new nature
  inspired method using a fuzzy combination of particle swarm optimization and
  genetic algorithms.
\newblock {\em Information Sciences}, 270:143--153, 2014.

\bibitem{vose-2017-genetic-algorithm}
A.~Vose, J.~Balma, G.~Wenes, and R.~Sukumar.
\newblock Deep neural network hyperparameter optimization with genetic
  algorithms.
\newblock 2017.

\bibitem{yi-2009-work-queue}
L.~Yi, C.~Moretti, S.~Emrich, K.~Judd, and D.~Thain.
\newblock Harnessing parallelism in multicore clusters with the all-pairs and
  wavefront abstractions.
\newblock In {\em Proceedings of the 18th ACM international symposium on High
  performance distributed computing}, pages 1--10. ACM, 2009.

\bibitem{zhang-2017-bot}
Y.~Zhang, J.~Sun, and Z.~Wu.
\newblock An heuristic for bag-of-tasks scheduling problems with resource
  demands and budget constraints to minimize makespan on hybrid clouds.
\newblock In {\em 2017 Fifth International Conference on Advanced Cloud and Big
  Data (CBD)}, pages 39--44, Aug 2017.

\bibitem{zhang-2016-bot}
Y.~Zhang, J.~Sun, and J.~Zhu.
\newblock An effective heuristic for due-date-constrained bag-of-tasks
  scheduling problem for total cost minimization on hybrid clouds.
\newblock In {\em 2016 International Conference on Progress in Informatics and
  Computing (PIC)}, pages 479--486, Dec 2016.

\bibitem{zoph-2016-archsearch}
B.~Zoph and Q.~V. Le.
\newblock Neural architecture search with reinforcement learning.
\newblock {\em arXiv preprint arXiv:1611.01578}, 2016.

\bibitem{zoph2017learning}
B.~Zoph, V.~Vasudevan, J.~Shlens, and Q.~V. Le.
\newblock Learning transferable architectures for scalable image recognition.
\newblock {\em arXiv preprint arXiv:1707.07012}, 2017.

\end{thebibliography}
}

\end{document}